\documentclass[conference]{IEEEtran}
\IEEEoverridecommandlockouts
\usepackage{cite}
\usepackage{amsmath,amssymb,amsfonts}
\usepackage{algorithm}
\usepackage[noend]{algpseudocode}
\usepackage{graphicx}
\usepackage{textcomp}
\usepackage{subcaption}
\usepackage{multirow}
\usepackage[table]{xcolor}
\def\BibTeX{{\rm B\kern-.05em{\sc i\kern-.025em b}\kern-.08em
    T\kern-.1667em\lower.7ex\hbox{E}\kern-.125emX}}
\begin{document}

\title{Segmentation and Defect Classification of the Power Line Insulators: A Deep Learning-based Approach\\}
\author{

\IEEEauthorblockN{Arman Alahyari, Anton Hinneck, Rahim Tariverdi, David Pozo}
\IEEEauthorblockA{{Center for Energy Science and Technology, } \\
{Skolkovo Institute of Science and Technology (\textit{Skoltech})}\\
}
}

\maketitle

\begin{abstract}
Power transmission networks physically connect the power generators to the electric consumers. Such systems extend over hundreds of kilometers. There are many components in the transmission infrastructure that require a proper inspection to guarantee flawless performance and reliable delivery, which, if done manually, can be very costly and time consuming. One essential component is the insulator. Its failure can cause an interruption of the entire transmission line or a widespread power failure. Automated fault detection could significantly decrease inspection time and related costs. Recently, several works have been proposed based on convolutional neural networks, which address the issue mentioned above. However, existing studies focus on a specific type of insulator faults. Thus, in this study, we introduce a two-stage model that segments insulators from their background to then classify their states based on four different categories, namely: healthy, broken, burned/corroded and missing cap. The test results show that the proposed approach can realize the effective
segmentation of insulators and achieve high accuracy in detecting several types of faults.   
\end{abstract}


\section{Introduction}

\subsection{Background and Motivation}
The smart grid is a concept that describes a novel structure of power grids in which smart communication and control facilities provide better reliability and resiliency towards failures. Network operation becomes more efficient by automation \cite{a1,ad3,ad1,ad2}.
Smart grids as a new design strategy for power grids consider three major sections; generation, transmission, and distribution. Many challenges exist towards full automation in all these three areas. However, as the transmission section can cover hundreds of kilometers, resolving related issues is of significant importance.


Even though transmission lines may seem to be the most important component, 
insulators play an important role as well. They mechanically secure wires in transmission sections. For instance, once any insulator turned faulty or defected, it directly influences the total usage and life span of transmission lines. Consequently, it is of great importance to detect insulator faults timely to guarantee the safety of transmission lines and improve the power grid's reliability \cite{a3}. 

\begin{figure}[!h]
\centering
  \includegraphics[width=\linewidth,trim={0cm 0 0 0cm},clip]{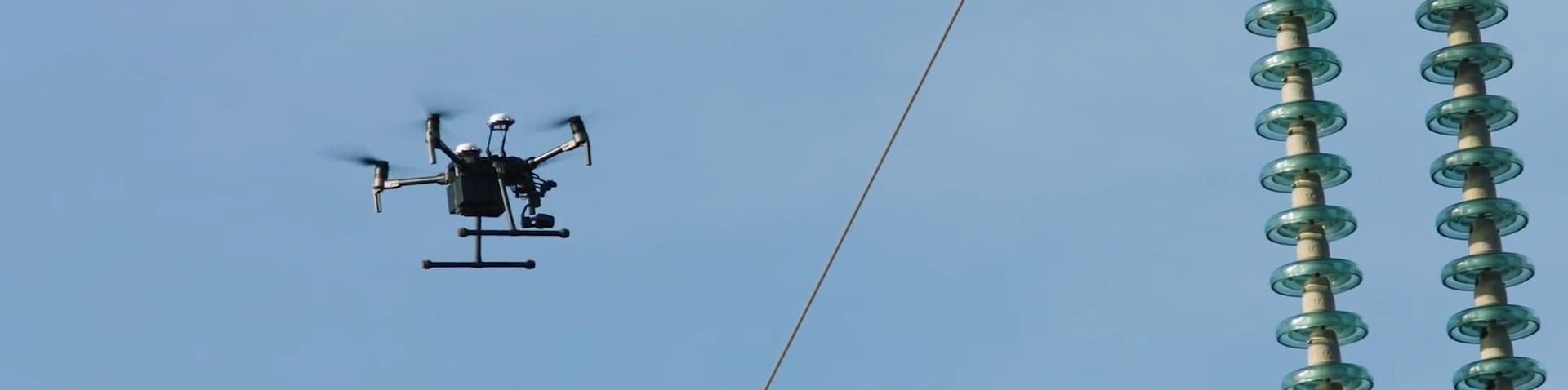}
  \caption{A UAV inspecting transmission line insulators \cite{china}.}
  \label{uav}
\end{figure}

Considering the length of transmission grids manual, regular inspections by helicopters and human experts, for example, are of great difficulty and due to the time and equipment required. Thus, several experiments were conducted exploring automated visual inspections of power transmission
systems \cite{a4,a5}. Specifically, considering the danger involved for humans, unmanned aerial vehicles (UAVs) have proven to be a useful tool in the inspection of power transmission grid components. Fig.\ref{uav} shows a UAV while it is inspecting the transmission lines insulators.

Images taken during the inspection by UAVs include different backgrounds such as mountains, fields, rural settlements or urban features. The lighting, including shadows and direct sunlight, varies dependent on the time of day and weather conditions (heavy rain, cloud coverage). Thus, automatically processing recorded footage is a complicated task.

\subsection{Literature Review and Contributions}
Two different practices exist to process insulators images, segmenting insulators and detecting defects. In recent years, several methods were proposed in studies on that topic \cite{a6,a7,a8,a9}. In  \cite{a6}, a detection approach for missing caps on transmission line insulators based on Faster Regions with a convolutional neural network is introduced. The given approach consists of a convolutional neural network (CNN) followed by a region proposal network and an object detector. Detection of the insulator images with the complex aerial background is also carried out in \cite{a7}. The model consists of five modules performing convolution and pooling and two modules of fully connected layers. The detection of insulators is done along with the identification of explosion faults. Similarly, in \cite{a8} a cascading CNN structure is proposed to perform segmentation and fault detection. The proposed network only detects missing caps. A transmission network insulator recognition system is investigated in \cite{a9}, where the segmentation algorithm is based on CNNs. In this study, the data-set is only utilized to segment and recognize insulators without any default detection.

\begin{figure}[ht]
\begin{subfigure}{.45\linewidth}
  \centering
  \includegraphics[width=.8\linewidth]{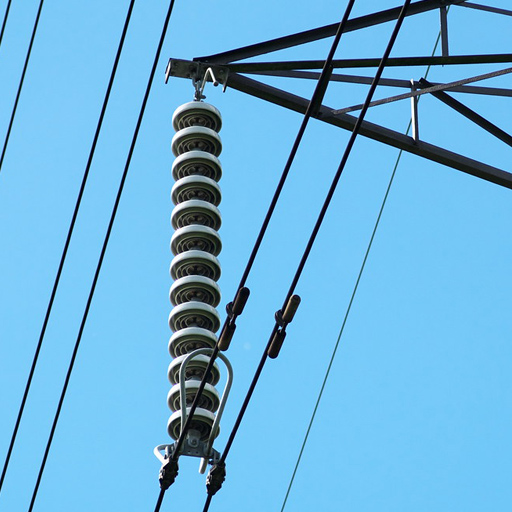}  
  \caption{Healthy}
  \label{fig:sub-first}
\end{subfigure}
\begin{subfigure}{.45\linewidth}
  \centering
  \includegraphics[width=.8\linewidth]{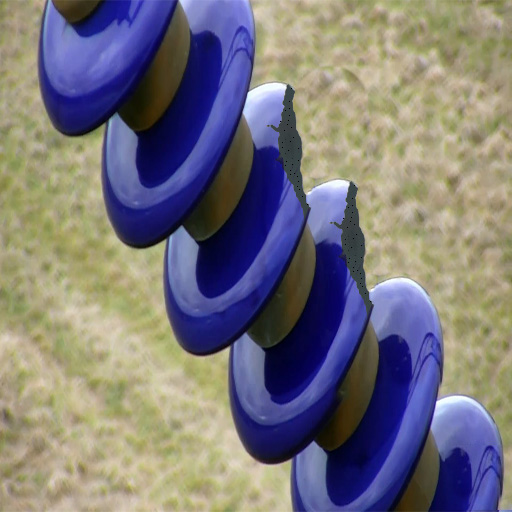}  
  \caption{Broken}
  \label{fig:sub-second}
\end{subfigure}\\
\begin{subfigure}{.45\linewidth}
  \centering
  \vspace{1ex}
  \includegraphics[width=.8\linewidth]{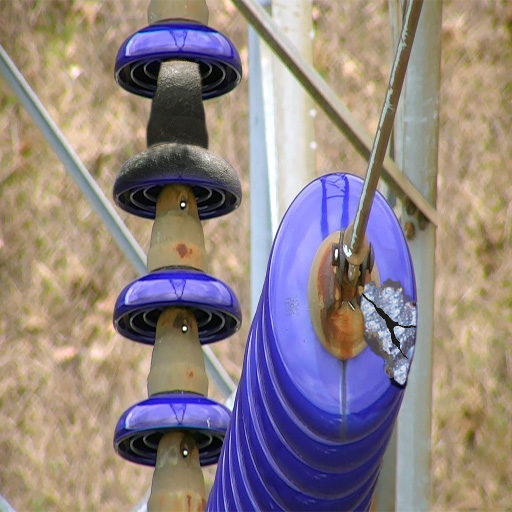}  
  \caption{Burned}
  \label{fig:sub-first}
\end{subfigure}
\begin{subfigure}{.45\linewidth}
  \centering
  \vspace{1ex}
  \includegraphics[width=.8\linewidth]{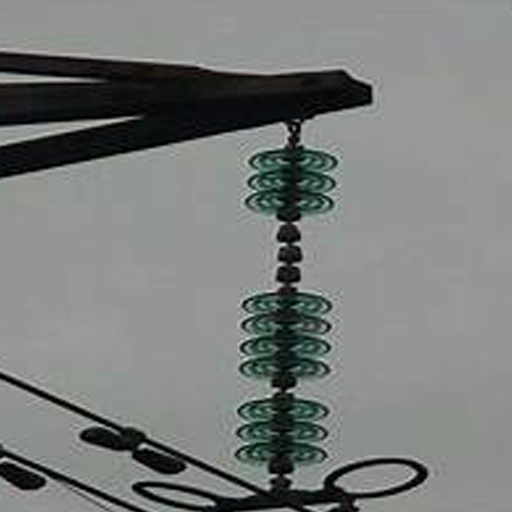}  
  \caption{Missing cap}
  \label{fig:sub-second}
\end{subfigure}
\caption{Modes of failure}
\label{fig:mof}
\end{figure}

The aforementioned studies either focus on segmentation tasks only or consider limited types of defects, for instance, missing caps. However, there are several failure modes documented for insulators. The most common failures are broken insulators, burn or corrosion damage and missing caps. Examples of such failures are shown in Fig.\ref{fig:mof}.

\subsection{Paper Contributions and Organization}
In this paper a two-stage model that involves both segmentation and detection tasks for faulty insulators detection is proposed. Both tasks are based on the use of state-of-the-art unsupervised learning methods. As a result, the defect detection can be fully automatized, given an image. We leverage the most recent developments on CNNs and develop an architecture and training method to achieve high-accuracy results. 
The algorithmic framework first segments the insulators in a taken images, to then classify their defects, if existent. In this work, we do not consider two classes only (e.g. healthy and defected), but classify a multitude of failure modes, which separates our work from studies carried out previously. Moreover, we use a wide range of different training and validation images.

The rest of the paper is organized as follows. Section \ref{sec.2} introduces the CNN architecture for segmentation and fault classification. Section \ref{sec.3} presents data preparation for the experiments. Experiments are described in section \ref{sec.4}. Finally, a conclusion is made in section \ref{sec.5}.

\section{Detection Approach}
\label{sec.2}
In this section, the utilized CNN structure and further details of network models, applied augmentations, loss functions and performance metrics are introduced.


\subsection{CNN Preliminaries}

CNN is a special type of neural network that has proven effective in computer vision applications. State-of-the-art results can be achieved in segmentation and classification tasks \cite{a10}. Compared to computer vision algorithms that do not take advantage of CNNs, much less pre-processing is required. More importantly, such networks are able to learn characteristics from data, which otherwise would have to be individually accounted for \cite{a11}.

Even though CNNs have been proposed in different architectures - to increase their efficiency for specific tasks or datasets, three different types of layers are used without exception, each with a specific purpose: convolutional, pooling, and fully connected (linear) layers. The \textit{convolutional layers} aim at extracting feature maps of the input images by applying filters over the different regions of images. For instance, with $k$ filters, each filter having weight and bias of $w_i$ and $b_i$, respectively, the convolution of an image patch, $x_n$, can be written as follows:

\begin{equation}
f_{i,n}=\sigma(W_ix_n+b_i),
\end{equation}

where $\sigma$ is the activation function. Besides rectified linear units (ReLU), sigmoid, or softmax activation functions, a multitude of different options exist, all having their individual advantages. These are applied on a layer's output neurons (e.g. after a convolutional layer).
After a number of convolutional layers, \textit{pooling layers} are commonly applied in prominent network architectures to reduce the size of particular dimensions. Max-pooling and average-pooling are two examples. Pooling layers, alongside reducing dimensions' sizes, perform denoising when utilized on images. 
\textit{Fully connected layers} are generally the last layers of CNNs, possessing a similar structure compared to the traditional neural networks\cite{a12}.

\subsection{Performance Metrics}
Let element $i \in \mathcal{I}$ denote a specific sample (image), which is subject to segmentation and classification. In the following, $\lvert \mathcal{I} \rvert = N^{\mathcal{I}}$ denotes the number of samples. The Intersection over Union (IoU) \eqref{eq2} is  used  for the  segmentation  problem (stage one in our framework). Let $\mathcal{P}_{i}$ be the set of all pixels in image $i$, with $N^{\mathcal{P}_{i}}$ denoting their total count. The ground truth value for pixel $p$ is denoted by $y_{p}$, whereas the prediction is denoted by $\hat{y}_{p}$. Since, a binary mask is predicted, every pixel's value $y_{p},\hat{y}_{p}\in \{0,1\}$. The IoU is computed based on the true positives ($\text{TP}_{i}$), true negatives ($\text{TN}_{i}$), false positives ($\text{FP}_{i}$) and false negatives ($\text{FN}_{i}$), $\forall~p \in \mathcal{P}_{i}$.
IoU is defined as 
\begin{equation}
\begin{split}
    &\text{IoU}_{i} = \frac{\text{TP}_{i}}{(\text{TP}_{i}+\text{FP}_{i}+\text{FN}_{i})}  = \\
    &= \frac{\sum_{p=1}^{N^{\mathcal{P}_{i}}} 1_{\left\{\hat{y}_{p}=y_{p}=1\right\}}}
    {\sum_{p=1}^{N^{\mathcal{P}_{i}}} 1_{\left\{\hat{y}_{p}=y_{p}=1\right\}} 
    + 1_{\left\{\hat{y}_{p}=1 ; y_{p}=0\right\}} 
    + 1_{\left\{\hat{y}_{p}=0 ; y_{p}=1\right\}}},
\end{split} \label{eq2}
\end{equation}
For the classification problem (second stage),  different metrics are used.
In multi-class classification a class $c\in\mathcal{C}$ is assigned to an image $i\in\mathcal{I}$.
Accordingly, $c_{i}$ denotes the ground truth label of an image, whereas $\hat{c}_{i}$ denotes its prediction. Insulators can either be healthy ($c=0$), broken ($c=1$), burned/corroded ($c=2$) or have caps missing ($c=3$).
The first metric, defined in equation \eqref{eq3}, is called accuracy (Acc).
\begin{equation}
\text{Acc} = \frac{\text{TP} + \text{TN}}{(\text{TP}+\text{TN}+\text{FP}+\text{FN})}  = \frac{1}{N^{\mathcal{I}}} \sum_{i=1}^{N^{\mathcal{I}}} 1_{\{\hat{c}_i=c_i\}} \label{eq3}
\end{equation}
Note, that TPs, TNs, FPs and FNs are now computed with respect to all classifications over all images, as opposed to an image's pixels.
For further analysis of the classification result, precision (Pcn), recall (Rcl) and F1 score (F1) metrics are utilized and defined in \eqref{eq4}--\eqref{eq4bis}, respectively.
\begin{IEEEeqnarray}{C}
\text{Pcn}_{c} = \frac{\text{TP}_{c}}{\text{TP}_{c} + \text{FP}_{c}}, \label{eq4}\\
\text{Rcl}_{c} = \frac{\text{TP}_{c}}{\text{TP}_{c} + \text{FN}_{c}}, \\
\text{F1}_{c} = 2*\frac{\text{Pcn}_{c} * \text{Rcl}_{c}}{\text{Pcn}_{c} + \text{Rcl}_{c}}\label{eq4bis}
\end{IEEEeqnarray}

As multi-class detection is utilized, these scores must be computed for each class individually. Results are displayed in section \ref{sec.4}.

\subsection{CNN Structure}
\begin{figure*}[!h]
 \centering
  \includegraphics[width = 0.7\linewidth]{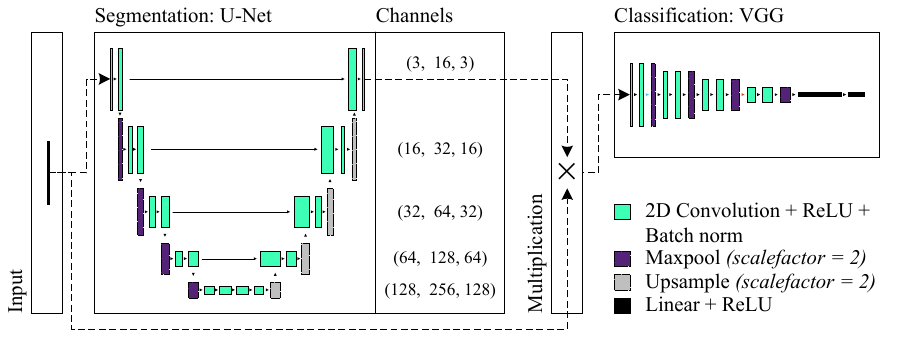}
  \caption{Structure of the fault detection algorithm}
  \label{fig:structure}
\end{figure*}

As discussed in the introduction section, the insulators' classification requires two different yet closely related stages. First, \textit{segmentation}  from the background image and only then apply the \textit{classification} network to determine their states (possible states are depicted in Fig.\ref{fig:mof}). 
Note that even though a single network may also be suggested to perform regarding the task mentioned above, in this paper, we propose a modular detection algorithm, as presented in  Fig.\ref{fig:structure} to deal with the problem in hand.

The structure includes two distinct yet well-known CNNs (see Fig.\ref{fig:structure}). The first network is utilized in the first stage segmenting the insulators, and the second network outputs the state of the insulator provided with the segmented insulator as input. Using two networks in serial allows for more straightforward implementation, as the segmentation and classification stages can be fine-tuned individually. Furthermore, a modular system facilitates replacing components in the data processing step. 

A CNN architecture that has proven effective for segmentation applications is the so-called UNet \cite{Ronneberger15}.
To classify the insulators' states, a VGG-like architecture is used \cite{Simonyan15}. First, a batch of input images is passed to the UNet. Its output is a batch of binary images, also known as as masks. Let $i$ denote an input image and $y$ denote the ground truth for the UNet's output. If pixel $p,~\text{with}~
p \in \mathcal{P}_{i}$ is part of an insulator, $y(p) = 1$, else $y(p) = 0$. Hence, $i\circ y$ yields an image with only the insulator(s) left. This reduces noise on the classification stage, because the image to classify only contains the component of interest.

\subsection{Loss Functions}

Both loss functions, the Weighted Binary Cross-Entropy (BCE), defined in equation \eqref{eq5}, as well as the Mean Squared Error (MSE), defined in equation \eqref{eq6}, are used for segmentation.
The predicted pixel values, as vectors, for a sample $i$ are denoted by $\hat{y}_{i}$ whereas the ground truth is denoted by $y_{i}$.
\begin{equation}
\begin{split}
\text{BCE} &=-\frac{1}{N} \sum_{i=1}^{N} w_{1} \cdot y_{i} \cdot \log \left(p\left(\hat{y_{i}}\right)\right) \\
& \quad + w_{2} \cdot \left(1-y_{i}\right) \cdot \log \left(1-p\left(\hat{y_{i}}\right)\right)
\end{split} \label{eq5}
\end{equation}

\begin{equation}
 \text{MSE} = \frac{1}{N}\sum_{i=1}^N \left\Vert y_i - \hat{y_i}\right\Vert^2  \label{eq6}
\end{equation}
 
In the scope of the second-stage classifier, BCE was chosen as the loss function as well. Note, that here, prediction and ground truth for a sample $i$ are scalar values, representing a specific class.
 
\section{Dataset Preparation}
\label{sec.3}
\begin{figure}[!h]
  \includegraphics[scale=0.3]{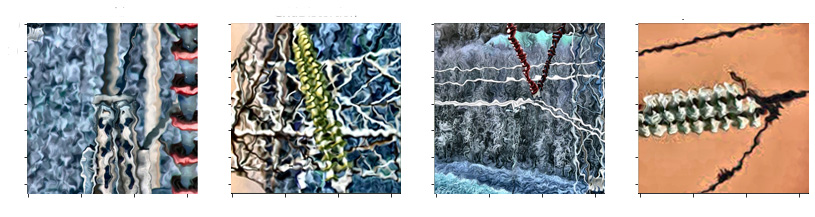}
\caption{Examples of the augmentation results}
\label{ris:aug_example1}
\end{figure}

In order to mimic the real-world behavior of how data is actually captured for analysis through the UAVs, we utilized directly the high-quality videos given by UAVs. These videos are taken by several companies throughout the world and are publicly accessible \cite{china,us}. This gives an advantage to our model. When training is completed, the network can recognize several types of insulators utilized in different locations and companies all over the world. We rendered the videos to image sequences frame by frame. Therefore, the input images are high quality images captured in these videos. In the next step of the preparation, we created ground truth for the segmentation task. The original number of insulators images was 119 however, with the augmentation tools, it was increased to 9520. To do so, we used different types of augmentations by albumentation \cite{albu} such as HorizontalFlip, VerticalFlip, BrightnessContrast, and others, complete list of which is given in Table \ref{tab:comp3}.  Also, some augmented images are shown in \mbox{Fig. \ref{ris:aug_example1}}.

We have three types of defects:  missing cap, broken cap, and burned cap samples of which are depicted in Fig. \ref{fig:mof}.  We also have separate types of augmentation applied for these images to have a sufficient number of images for this task. The number original images and augmented are specified in Table \ref{tab:table-name}.

\begin{table}
\caption{Number of insulator images before augmentation}
\label{tab:table-name}
\begin{center}
\begin{tabular}{ |c|c|c| } 
\hline
Data & Classification & Segmentation  \\
\hline
\multicolumn{3}{|c|}{Before augmentation}\\
\hline
\multirow{1}{*}{Train-set} & 46 & 119  \\ 
\multirow{1}{*}{Validation-set} & 14 & 14 \\ 
\hline
\multicolumn{3}{|c|}{After augmentation}\\
\hline
\multirow{1}{*}{Train-set} & 3680 & 9520  \\ 
\multirow{1}{*}{Validation-set} & 14 & 14 \\
\hline
\end{tabular}
\end{center}
\end{table}

\begin{table}[]
    \caption{Augmentation Table}
\centering
    \small
        \begin{tabular}{ |p{3cm}|p{1.5cm}|  }
        \hline
        Augmentation Types & Probabilities  \\
        \hline
        VerticalFlip & p=0.5 \\
        HorizontalFlip & p=0.5 \\
        ElasticTransform & p=0.5 \\
        GridDistortion & p=0.5  \\
        OpticalDistortion & p=0.8  \\
        Transpose & - \\
        RandomRotate90 & p=0.5  \\
        CLAHE & p=0.8 \\
        RandomBrightness &p=0.5\\
        RandomContrast&p=0.5\\
        RandomGamma & p=0.8 \\
        \hline
        \end{tabular}
    \label{tab:comp3}

\end{table}

\section{Experiments} 
\label{sec.4}
As mentioned in the previous section, the structure of the CNN to detect faults includes two different CNNs.
As shown in the illustration, the output of the UNet (a binary mask) is passed to the VGG, alongside the input image, for classification. Thus, the classification returned by the VGG is dependent on the segmentation result of the UNet.
Training UNet is straight forward as it can be trained by considering the available  ground truth of the target masks.
However, for training the classification network two approaches can be investigated:
\begin{itemize}
    \item Training UNet and VGG independently
    \item Training VGG on UNet's (trained) output
\end{itemize}
In the remainder of this subsection, we first investigate UNet (segmentation) training  and its performances and then the result of both approaches for classification stage training are discussed.
\subsubsection{UNet training} 

\begin{figure}[ht]
\begin{subfigure}{\linewidth}
  \centering
  \includegraphics[width=.9\linewidth]{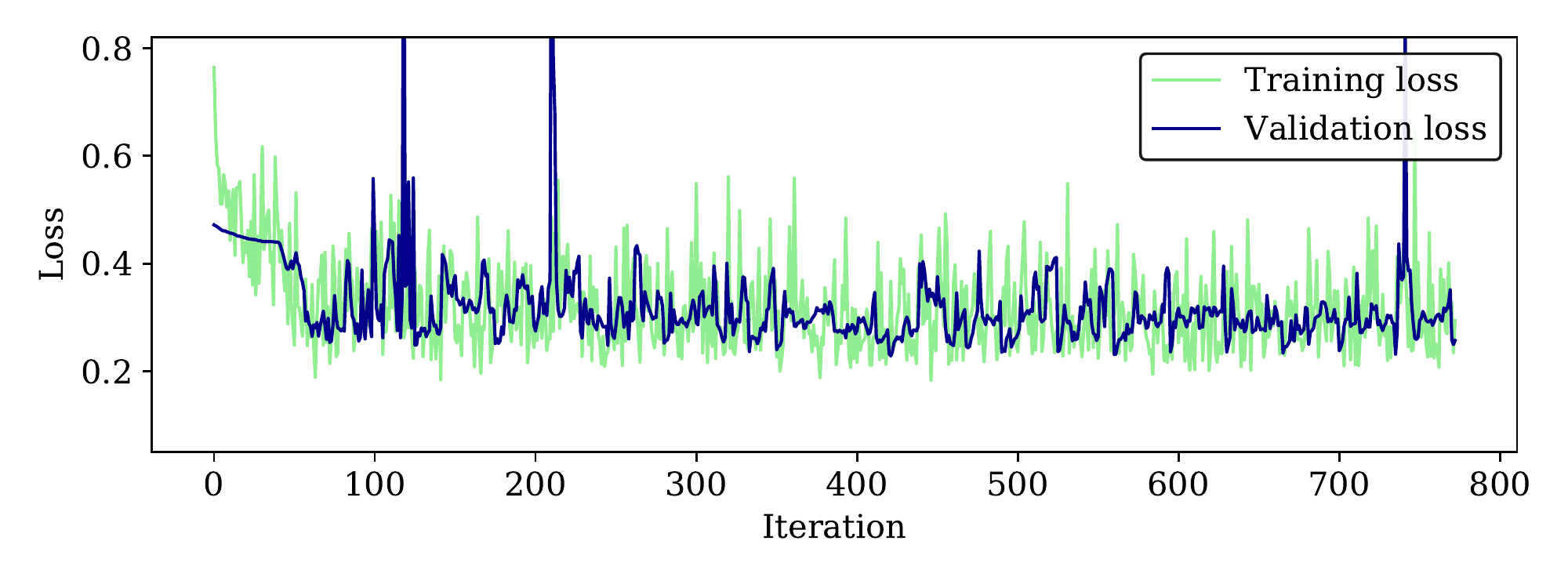}
  \caption{Training sequence 1}
  \label{fig:vggtrain1}
\end{subfigure}\\
\begin{subfigure}{\linewidth}
  \centering
  \vspace{1ex}
  \includegraphics[width=
  .9\linewidth]{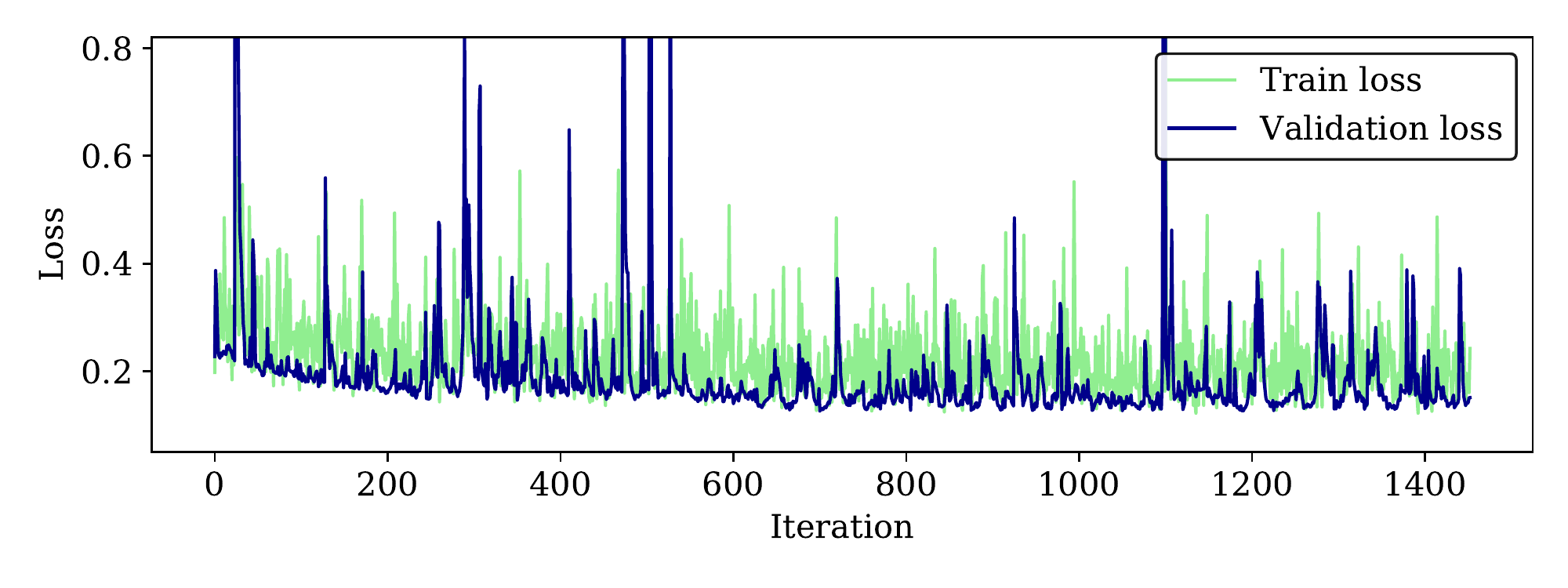}  
  \caption{Training sequence 2}
  \label{fig:vggtrain1}
\end{subfigure}
\caption{UNet training}
\label{fig:unettrain}
\end{figure}

\begin{figure}[ht]
	\center{\includegraphics[width=.9\linewidth]{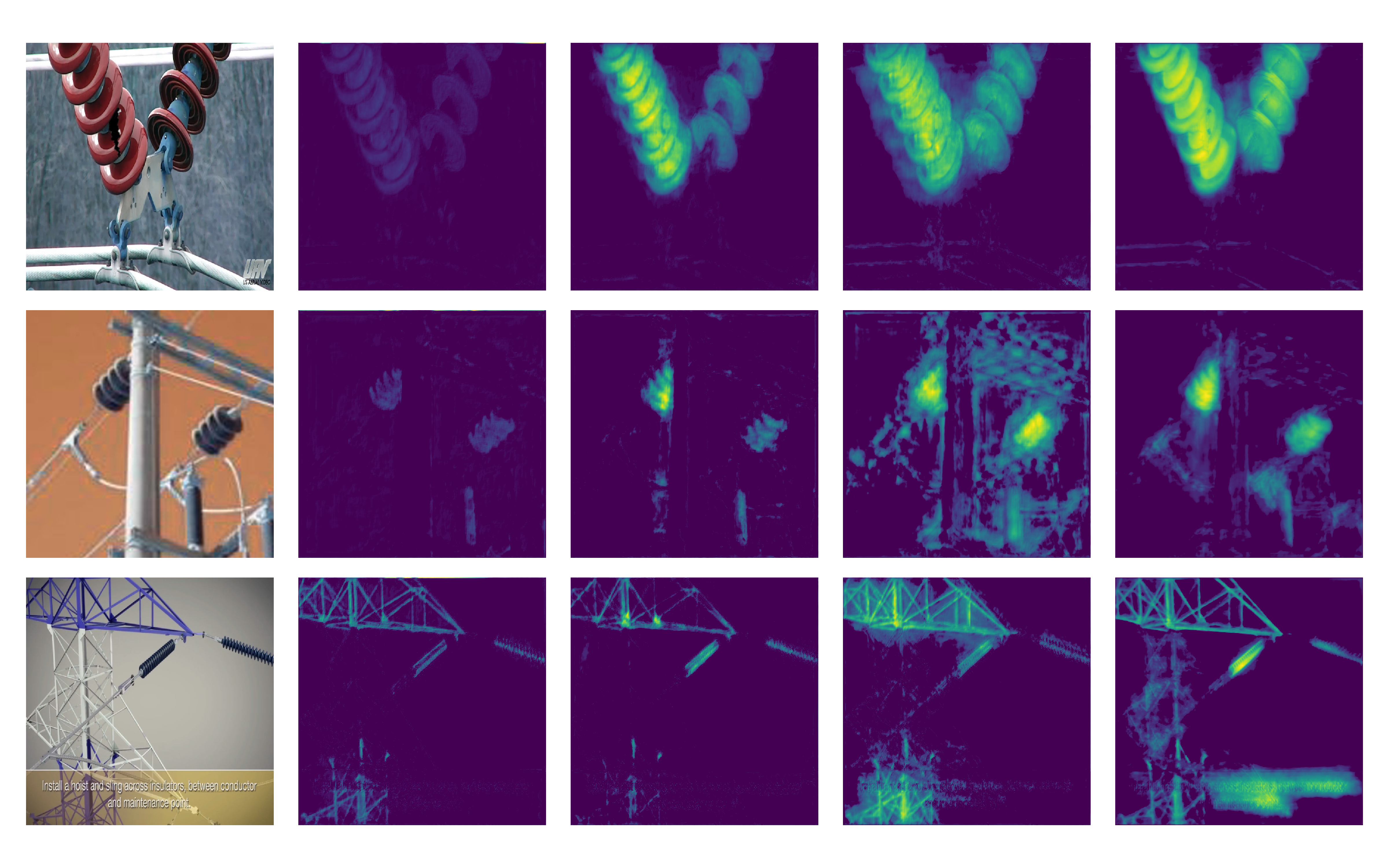}}
	\caption{Output of the proposed UNet architecture at different stages in the training process (100, 500, 1200 and 2200 iterations)}
	\label{fig:iou}
\end{figure}

\begin{figure}[ht]
	\center{\includegraphics[width=.9\linewidth]{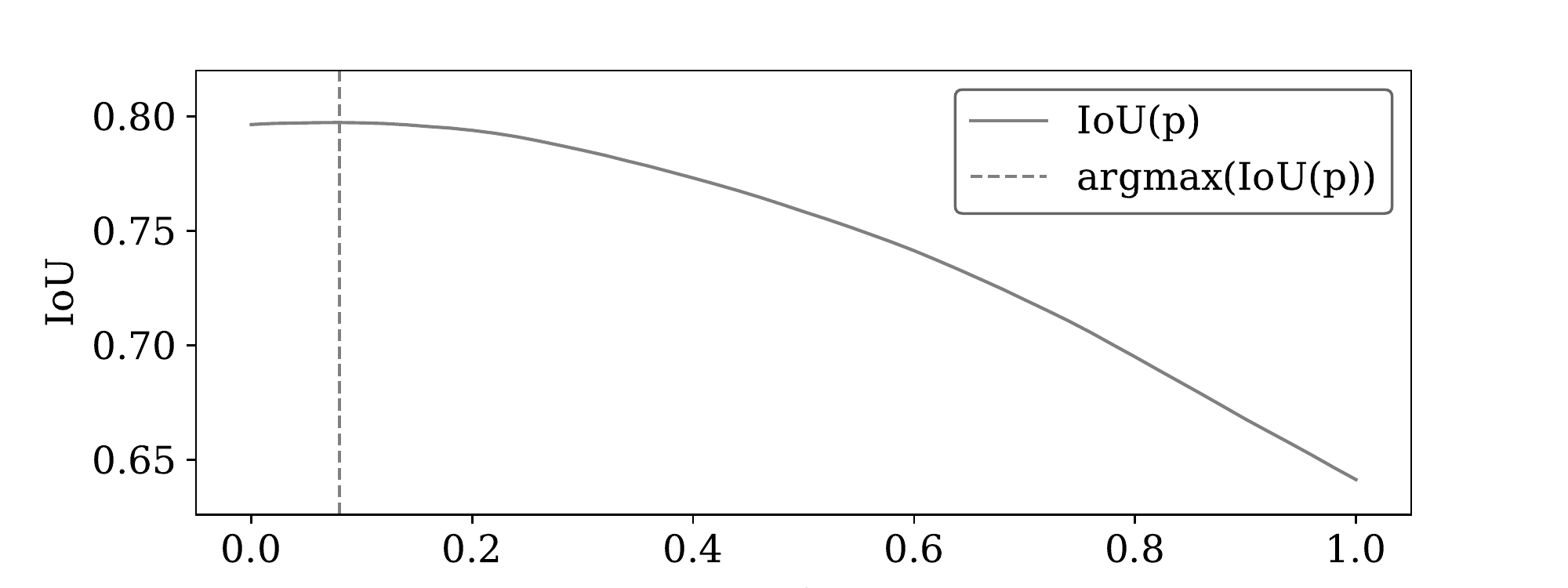}}
	\caption{IoU as a function of thresholds $p$}
	\label{fig:iou1}
\end{figure}

\begin{figure}[!h]
	\center{\includegraphics[width=.8\linewidth]{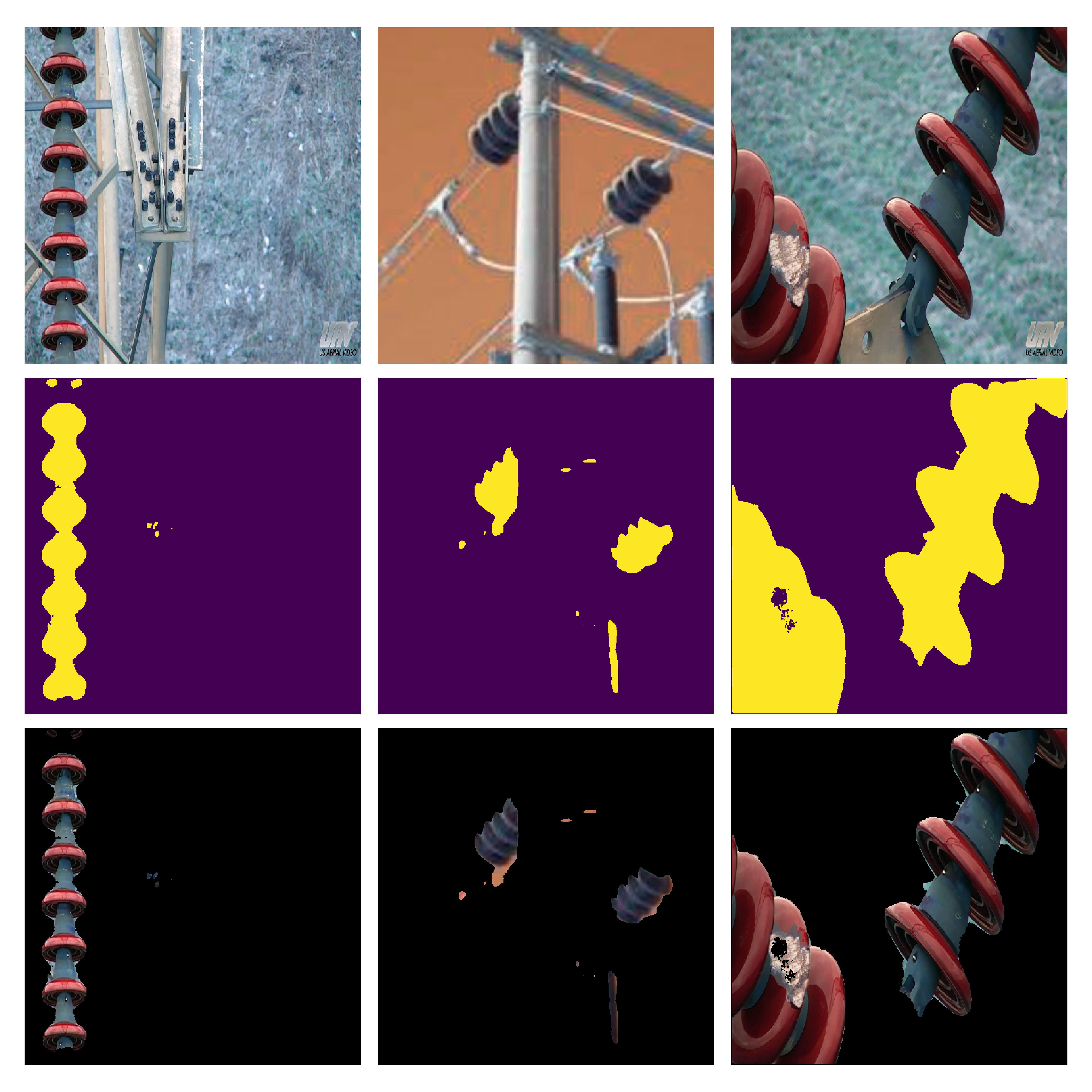}}
	\caption{Segmentation results: original image (top), the generated mask (middle), segmented image (bottom)}
	\label{fig:seg}
\end{figure}

Since, the original dataset is captured from several videos with different background and number of the images is rather low, we propose training to be carried in two sequence which guarantees a better training result and performance in the end. These two sequences are depicted in \mbox{Fig. \ref{fig:unettrain}}. As can be seen from this figure in the second sequence lower losses are achieved for the validation set. Note that the depicted loss is calculated over batch thereby including some fluctuation. However, only the best performance is saved and utilized in the next steps. Several stage of the training and the related outputs are depicted in \mbox{Fig. \ref{fig:iou}}. We also investigated a threshold where UNet can preform its best performance considering maximum IoU metric as shown in \mbox{Fig. \ref{fig:iou1}}. The Trained segmentation network could reach the IoU of 0.795 in segmenting the insulators from the main images. Some of the final segmentation network results and produced masks are depicted in the \mbox{Fig. \ref{fig:seg}}. As can be seen, the trained UNet  performs satisfactory in segmentation of power transmission insulators. 

\subsubsection{VGG}

\begin{figure}[h!]
	\center{\includegraphics[width=\linewidth]{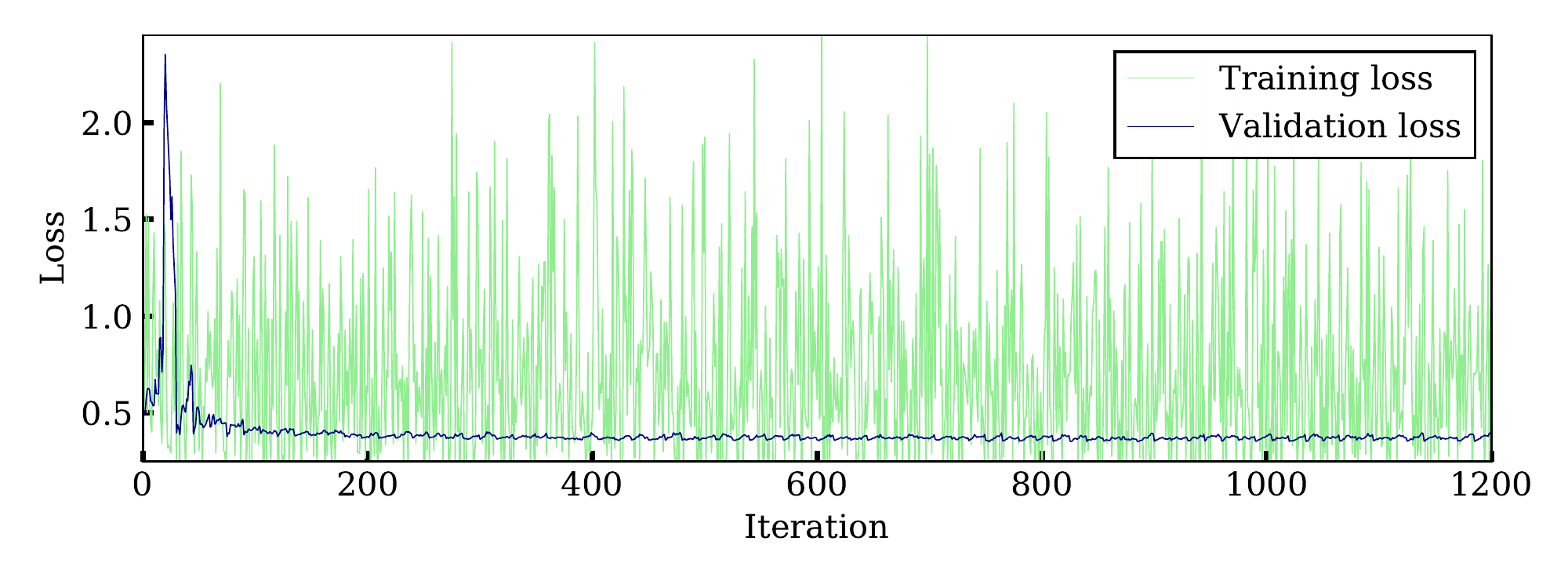}}
	\caption{VGG training on correct ground truth}
	\label{fig:vggtrain}
\end{figure}

Since the prediction for the proposed network structure is computed as $\hat{y} = \mathrm{VGG}(\mathrm{U\text{-}Net}(x), x)$, the classification result is dependent on the output of the segmenting unit. Two different types of training were conducted. First, the VGG was trained on the ground truth segmentation mask. The obtained loss through the explained regime is shown in \mbox{Fig. \ref{fig:vggtrain}}. Stochastic gradient descent was chosen as an optimizer method, with a learning rate of 0.008 and a momentum of 0.26. Furthermore, a multiplicative scheduler was employed multiplying the learning rate by 0.98 after every global step. With the exogenous ground truth the obtained classification network accuracy is 0.92.

As shown in Table \ref{tab:alg1} however, utilizing both independently trained networks yields a mere 21\% accuracy. This suggests that classification, as outlined, highly depends on the result of segmentation network. Training these components independently leads to poor outcomes. 
Secondly, the classifier was trained on segmentation results produced by the trained \mbox{U-Net}. 

Improved results were obtained by training the classifier on the output of the segmentation stage, as specifics proceeding of CNN could be considered. Note that getting acceptable training result proved difficult, due to limitations (especially in size) of the created dataset. Moreover, the dataset is very diverse. Thus, several training approaches were investigated.
\subsubsection{Pre-training}
CNNs are non-convex functions. The training success can be highly dependent on the starting point of the optimization. Rather than initializing the network randomly, a starting point can be supplied by loading model weights. In the following, pre-training means setting the model's weights to the result obtained by training the VGG on ground truth data. This had positive effects on the consecutive training on segmentation results.
\subsubsection{Reset and Alteration}

The training was conducted in two nested for-loops, looping over outer and inner epochs. After looping over all inner epochs, adjustments were made to the optimizer.
The first technique that proved successful was resetting the solver to its initial state. This prevented diminishing improvements when a local minimum was reached, or the learning rate became too small. This regularly facilitated finding an even better solution after a reset.
Secondly, it was tried alternating between optimizers with different settings, in the same manner. This technique, however, turned out to be less effective.

\begin{figure}[h!]
\begin{subfigure}{\linewidth}
  \centering
  \vspace{1ex}
  \includegraphics[width=
  .9\linewidth]{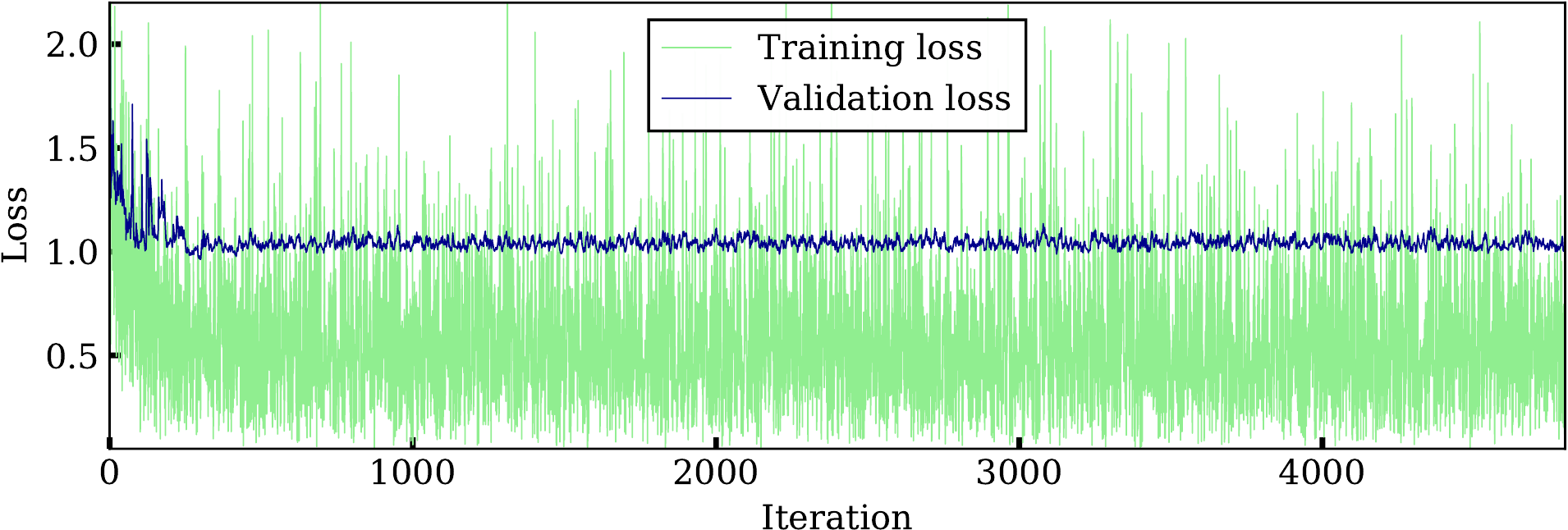}  
  \caption{No pre-training and no solver reset \\\vspace{0.5cm}}
  \label{fig:vggtrain1} 
\end{subfigure}\\
\begin{subfigure}{\linewidth}
  \centering
  \includegraphics[width=.9\linewidth]{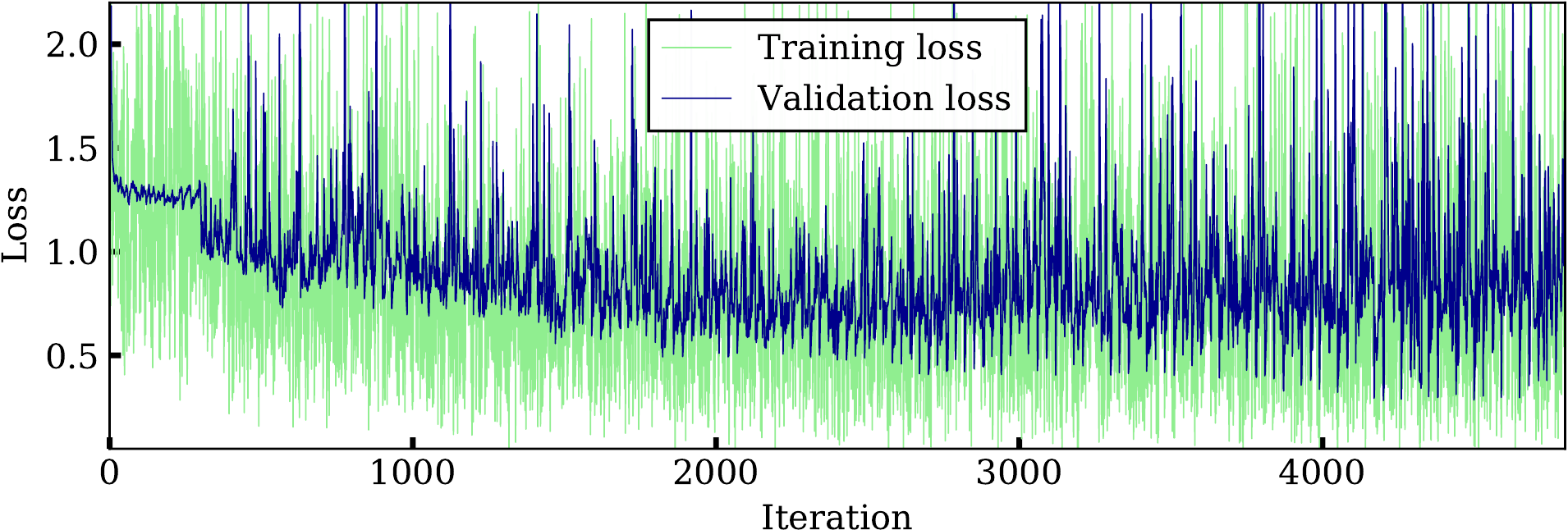}
  \caption{Pre-training and solver reset \\\vspace{0.5cm}}
  \label{fig:vggtrain0}
\end{subfigure}\\
\caption{VGG training on UNet}
\label{fig:vgg_unet}
\end{figure}

 For instance, the obtained loss values in training are depicted in \mbox{Fig. \ref{fig:vggtrain1}} where no pre-training or solver resetting is utilized. It is clear increasing training iteration has no effect in decreasing validation loss. While applying these two, as shown in Fig.\ref{fig:vggtrain0} could lead to much better results as the minimum validation loss gradually decreases over training iterations. 
 
Finally, the acquired accuracy by the combination of the aforementioned approaches is concluded in Table \ref{tab:alg1}. As can be observed in \mbox{Fig. \ref{fig:mof}}, the data set's pictures are fairly diverse meaning that they include several types of insulators with dissimilar backgrounds, that is why the classification accuracy is limited to 0.78. However, if our proposed approach including the utilization of CNN and training approach is utilized by a specific transmission maintenance company, the diversity is noticeably reduced. Thus, we also evaluated the proposed approach on a  more homogeneous data set. We ran 20 training epochs initializing the network with previous weights and obtained  an accuracy of 93\% as also reported in Table \ref{tab:alg1} which is quite satisfactory for the aforementioned classification problem.

\begin{table}[!htb]
	\caption{The obtained accuracy results for different training regime of VGG based on UNet outputs (4800 global iterations)}
	\label{tab:alg1}
	\begin{center}
		\small
		\begin{tabular}{| c | c  c  c | c  |}
		    \hline
			\# Training & Pre-trained & Reset & Alternating & $\text{Acc}$ \\
			\hline
			1  & x & - & - & 0.42\\
			2  & - & x & - & 0.5\\
			3  & x & - & x & 0.71\\
			4  & x & x & - & \textbf{0.78}\\
			\hline
			\multicolumn{4}{|c|}{\textit{Performance on homogeneous data set}} & \textbf{0.93}\\
			\hline
		    \multicolumn{4}{|c|}{\textit{Separated VGG and UNet training}} & 0.21\\
		    \hline
		\end{tabular}
	\end{center}
\end{table}

\begin{table}[!htb]
	\caption{Additional performance metrics}
	\label{tab:metrics}
	\begin{center}
		\small
		\begin{tabular}{ | c || c c c c | }
		    \multicolumn{1}{c}{} & \multicolumn{4}{c}{Classes$^a$}\\
		    \hline
			Metric & 0 & 1 & 2 & 3 \\
			\hline
			\text{Pcn} & 0.6 & 1.0 & 1.0 & 0.67\\
			\text{Rcl} & 0.75 & 0.75 & 0.75 & 1.0\\
			\text{F1} & 0.67 & 0.86 & 0.86 & 0.8\\
			\hline
		\end{tabular}
	\end{center}
	\footnotesize{$^a$ $c=0$: healthy, $c=1$: broken, $c=2$: burned/corroded, $c=3$: missing caps}
\end{table}

Table \ref{tab:metrics} gives additional insights. Precision, recall and consequently F1 score are the lowest for $c = 0$. Classes $c = 1$ and $c = 2$ show equally good classification scores. One reason might be, that burned or corroded insulators not only show specific geometric properties but information in the form of colors as well. The same can apply to broken insulators. An implication might be, that using automated methods better image quality (and therefore a larger amount of retained information) is highly advantageous.

\section{Conclusion}
\label{sec.5}
Insulators  are known to be a vital component of the
energy transmission systems. Yet, they are exposed to electrical, mechanical, and environmental stresses leading to different kinds of defects. In this study, we focused on investigating approaches based on the deep learning techniques to detect and classify defected insulators. In doing so, we utilized state-of-art CNNs such as UNets and VGG modified to match the task in hand. We had three different defects. Namely, insulators with broken, burned, and missing caps. The results showed, despite the lack of a large input dataset, good performance of the trained UNet in the segmentation task of insulators and the trained VGG as the second stage network in reaching a high IoU and a high accuracy classifying different damage states. These experiments could be extended using a larger amount of aerial imagery and different CNN architectures.

\bibliographystyle{IEEEtran}
\bibliography{ref.bib}
\end{document}